\documentclass[10pt,twocolumn,letterpaper]{article}

\usepackage{cvpr}

\usepackage{booktabs}
\usepackage{multirow}
\usepackage{amsmath,amssymb}
\usepackage{microtype}
\usepackage{enumitem}
\usepackage{xcolor}
\usepackage{graphicx}
\usepackage{ifthen}
\usepackage{dblfloatfix}

\definecolor{projectblue}{RGB}{0,82,155}
\definecolor{cvprblue}{rgb}{0.21,0.49,0.74}
\usepackage[pagebackref,breaklinks,colorlinks,allcolors=cvprblue]{hyperref}

\def\paperID{0000}
\def\confName{CVPR}
\def\confYear{2026}

\title{
\Large\bfseries
WireSeg-32K: \\A Physics-Grounded Synthetic Dataset for Wire Instance Segmentation
\\[-0.15em]
{\small\normalfont\sffamily\bfseries
\href{https://deformx.github.io/}{
  \textcolor{projectblue}{Project Page: deformx.github.io}
}}
\vspace{-1.2em}
}

\author{
Zilin Dai\textsuperscript{1} \quad
Lehong Wang\textsuperscript{2,\dag} \quad
Yi Yang\textsuperscript{2,\dag} \quad
Xiang Fei\textsuperscript{2,\dag}
\\[0.35em]
\textsuperscript{1}Harvard University \qquad
\textsuperscript{2}Carnegie Mellon University
\\[0.2em]
{\tt\small zilin\_dai@g.harvard.edu \quad
\{lehongw, yiyang, xiangfei\}@andrew.cmu.edu}
}

\setlist[itemize]{leftmargin=*,topsep=2pt,itemsep=1pt,parsep=0pt}
\setlist[enumerate]{leftmargin=*,topsep=2pt,itemsep=1pt,parsep=0pt}

\newcommand{\safeincludegraphics}[2][]{%
  \IfFileExists{#2}{%
    \includegraphics[#1]{#2}%
  }{%
    \fbox{\parbox[c][0.18\textheight][c]{0.95\linewidth}{\centering
      Placeholder for original figure\\[2pt]\texttt{\detokenize{#2}}}}%
  }%
}

\newcommand{\Yes}{\checkmark}
\newcommand{\No}{\ensuremath{\times}}

\begin{document}
\maketitle
\renewcommand{\thefootnote}{}
\footnotetext{\textsuperscript{\dag}Corresponding authors and project leads.}
\renewcommand{\thefootnote}{\arabic{footnote}}

\begin{abstract}
Deformable linear objects such as wires and cables are difficult to segment because they are thin, highly deformable, and frequently self-occluded, while large-scale instance-level annotations are expensive to obtain in real scenes. Existing resources either focus on cable tracing or semantic segmentation under constrained settings, or generate visually plausible images without physically grounded wire deformation. We present \textit{WireSeg-32k}, a synthetic dataset for wire instance segmentation with 32,000 RGB images, instance masks, depth maps, and a complementary real-world test set with annotations. To generate this dataset, we develop \textit{DeformX}, a co-simulation pipeline that couples Cosserat-rod dynamics with photorealistic Isaac Sim rendering, enabling physically plausible, contact-consistent wire shapes, CAD-based wire assets, and diverse visually grounded scenes. As a simple baseline, LoRA fine-tuning SAM3 on WireSeg-32k alone improves real-world mAP@75 by 10.2\% over the off-the-shelf model, showing that physically grounded synthetic data can transfer to real wire perception.
\end{abstract}

\vspace{-10pt}
\section{INTRODUCTION}

Deformable linear objects (DLOs) such as wires, cables, and ropes are common in robotics and industrial environments, including wire harness assembly, cable routing, maintenance, and inspection~\cite{intro_survey,wang2024wireharnessreview}. Reliable visual perception is a prerequisite for these tasks, yet wire instance segmentation remains difficult: wires are thin, visually repetitive, frequently occluded, and expensive to annotate at scale in cluttered scenes. As a result, large real-world datasets with instance-level wire labels remain scarce, making synthetic data an attractive path toward scalable supervision.

Existing datasets and synthetic pipelines only partially address this need. HANDLOOM~\cite{handloom} provides large-scale synthetic data for cable tracing, but its supervision is trace-based and its setting focuses on grayscale, semi-planar cable images. Fresnillo et al.~\cite{fresnillo2024} generate 25k photorealistic cable images that are effective for semantic segmentation, but do not provide per-wire instance masks. Earlier auto-generated wire datasets~\cite{zanella2021} and perception methods such as FASTDLO~\cite{fastdlo2022} and ISCUTE~\cite{iscute2024} have advanced DLO segmentation, but the data landscape remains fragmented across semantic masks, cable traces, or constrained scene assumptions. For a dataset aimed at modern wire perception, we need instance-level labels, visually grounded scenes, and wire configurations that are not only diverse in appearance but also plausible under gravity, contact, and self-interaction.

In this work, we center on \textit{WireSeg-32k}, a synthetic dataset designed to fill this gap. WireSeg-32k contains 32,000 rendered images from 300+ simulation runs across tabletop, hanging-wire, and data-center scenarios, together with per-wire instance masks, dense depth maps, and easy/medium/hard splits. To generate it, we develop \textit{DeformX} \cite{yang2026deformx}, a co-simulation framework combining a Cosserat rod engine with Isaac Sim so that dataset generation benefits from both physically grounded DLO dynamics and photorealistic rendering. DeformX is therefore the tool that enables the dataset, rather than the final end product of the paper.

We make three contributions:
\begin{itemize}
    \item \textbf{WireSeg-32k:} A 32k-image synthetic wire instance segmentation dataset with per-wire masks, depth maps, difficulty splits, and an annotated real-world test set.
    \item \textbf{DeformX for dataset generation:} A co-simulation pipeline for generating physically plausible and visually realistic wire data using Cosserat-rod dynamics, CAD-compatible assets, and photorealistic rendering.
    \item \textbf{A simple transfer baseline:} A lightweight LoRA fine-tuning recipe on SAM3 trained only on WireSeg-32k that improves real-world mAP@75 by 10.2\%, demonstrating sim-to-real value even without real-image training.
\end{itemize}

\begin{figure*}[t]
    \centering
    \safeincludegraphics[width=\textwidth]{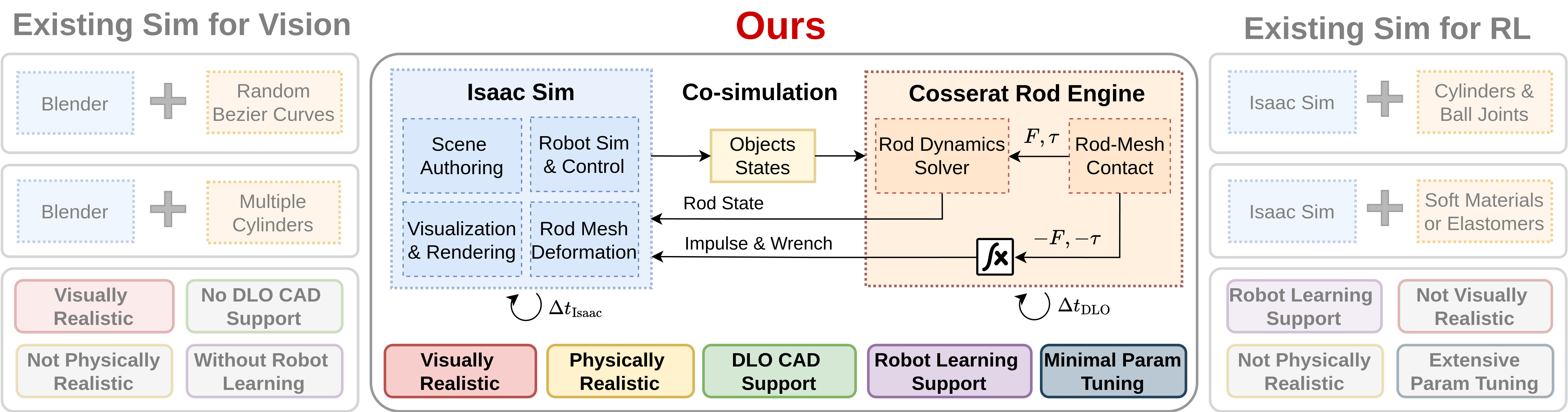}
    \caption{Overview of our co-simulation framework. Compared to existing simulators for vision (left) and RL (right), our framework jointly achieves visual realism, physical accuracy, DLO CAD support, and robot learning support with minimal parameter tuning by combining Isaac Sim and a dedicated Cosserat rod engine.}
    \label{fig:system}
\end{figure*}

\setcounter{figure}{1}
\begin{figure}[h]
    \centering
    \includegraphics[width=0.8\linewidth]{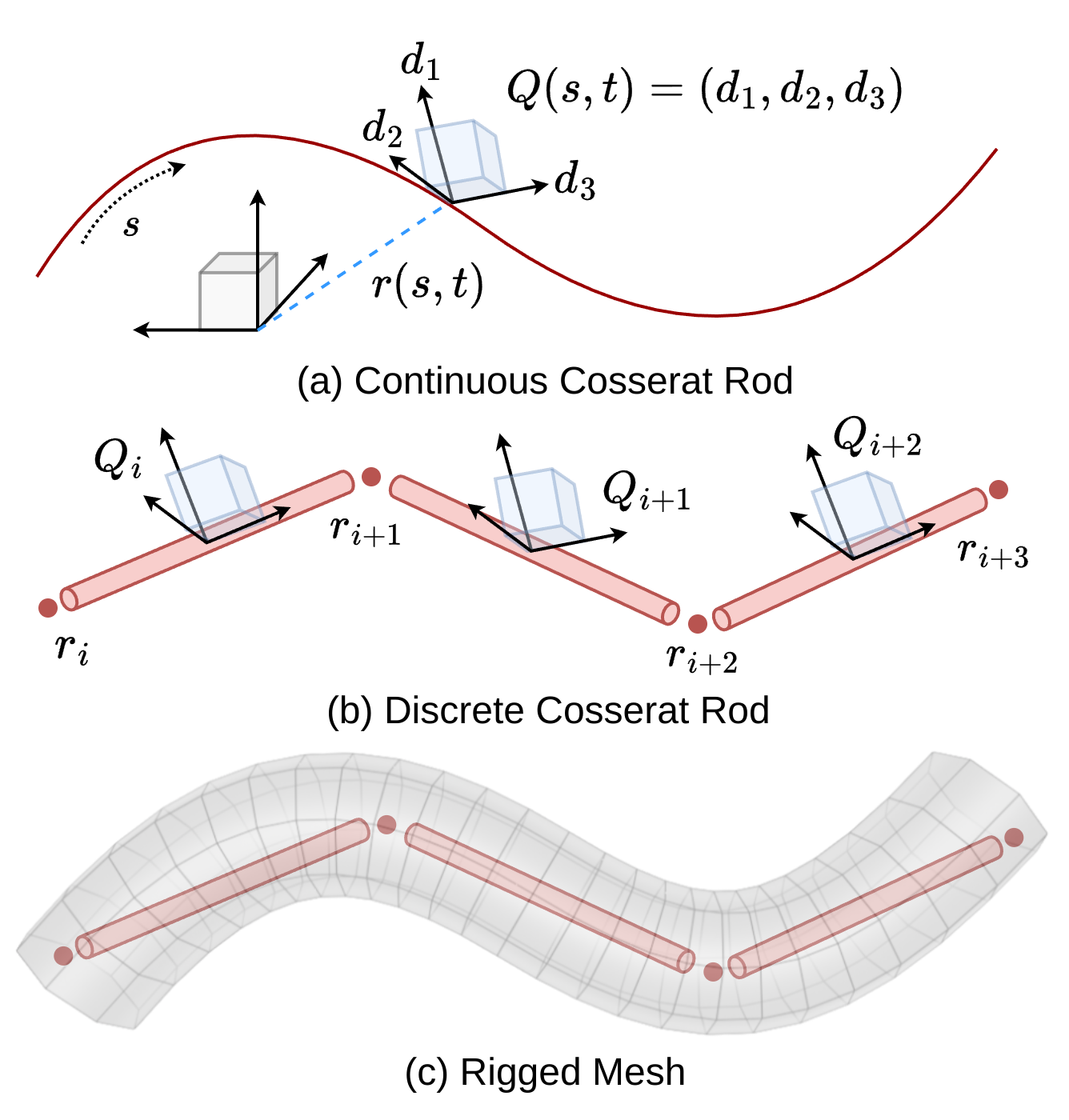}
    \caption{Cosserat rod modeling of DLOs. (a) Continuous Cosserat rod representation; (b) Discrete Cosserat rod representation; (c) Mesh skinned to the discrete Cosserat rod for realistic visualization.}
    \label{fig:notation}
\end{figure}

\section{Method: DeformX}
\noindent\textbf{Physics model.}
To generate realistic wire configurations, DeformX models each DLO as a Cosserat rod, a slender-body continuum representation that captures stretching, shearing, bending, and twisting~\cite{pyelastica2021}. In the continuous view, a rod is parameterized by a centerline $\mathbf{r}(s,t)\in\mathbb{R}^3$ and an orthonormal material frame $\mathbf{Q}(s,t)\in SO(3)$ along arc length $s$. In practice, we discretize the rod into vertices and segment frames, and integrate its dynamics in a dedicated rod engine. Compared with procedural curves or rigid-link approximations commonly used in synthetic wire generation~\cite{zanella2021,fresnillo2024}, this gives physically interpretable control over gravity-driven sagging, bending stiffness, torsion, and quasi-static shape transitions. These properties matter for data generation because the visual ambiguity of wire instance segmentation is often created by physically plausible crossings, draping, and self-occlusion rather than by arbitrary curve sampling alone.

\begin{figure}[t]
    \centering
    \safeincludegraphics[width=\linewidth,height=0.46\textheight,keepaspectratio]{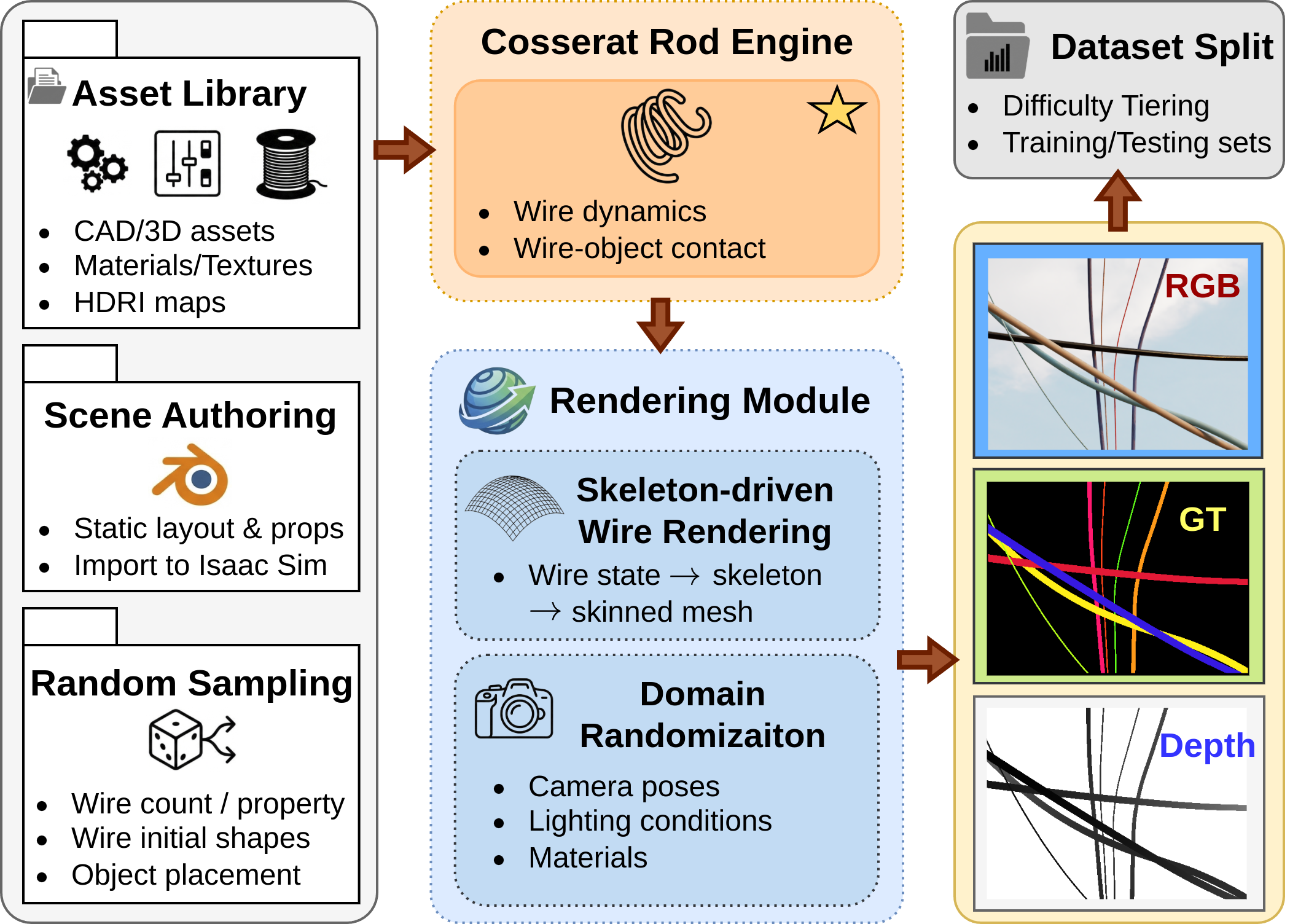}
    \caption{Synthetic wire segmentation dataset generation pipeline. Randomized scenes are built from asset libraries. Wires are simulated with a Cosserat rod engine, and visualized as skinned meshes. RGB images, segmentation masks, and depth maps are generated and split by difficulty levels.}
    \label{fig:data_pipeline}
\end{figure}

\noindent\textbf{Free-form mesh contact.}
Realistic wire data requires interaction with cluttered geometry. While existing Cosserat rod simulators provide strong mechanics, they do not natively support contact-rich synthetic scenes with arbitrary CAD assets. DeformX therefore adds free-form mesh contact instead of limiting interactions to planes or analytic primitives. Building on a penalty-based formulation, we compute closest-point distances between rod nodes and mesh surfaces, then apply repulsive normal forces with stiffness and damping, while returning reaction impulses to the corresponding rigid objects. To accelerate contact queries, meshes are organized with broad-phase pruning, and a repulsion distance is introduced to reduce deep penetration at larger time steps. This allows wires to drape over structures, slide along supports, wrap around obstacles, and form realistic multi-point contacts common in cable scenes.

\noindent\textbf{Mesh skinning and co-simulation.}
The rod state is converted into a renderable wire through skeleton-driven mesh skinning: a smooth tubular or CAD-derived mesh is bound to the simulated rod vertices and deformed according to the rod state. This preserves physical consistency while producing visually realistic cable geometry compatible with asset libraries and imported DLO meshes. DeformX couples the rod engine to Isaac Sim through a lightweight multi-rate interface: the rod solver advances the DLO at fine substeps, while Isaac Sim updates rigid bodies, cameras, lighting, and rendering at a coarser rate. This lets users manipulate physically realistic DLOs in photorealistic scenes while relying on a dedicated rod solver underneath. The same interface supports both interactive scene authoring and large-scale headless data generation.

\noindent\textbf{Data-generation pipeline.}
Figure~\ref{fig:data_pipeline} summarizes the pipeline used to build WireSeg-32K. We first assemble scene templates from publicly available assets and Isaac Sim environments, then sample scene objects, textures, camera viewpoints, and lighting conditions. For each scene instance, we generate wire configurations by varying the number of wires, lengths, radii, initial shapes, and material parameters, and simulate them with DeformX while the rest of the scene is managed by Isaac Sim. The resulting rod states drive skinned meshes for rendering, and because labels remain tied to the simulated DLO identities throughout the pipeline, we can export RGB images, per-wire instance masks, and depth maps consistently even under heavy deformation, occlusion, and clutter. In short, DeformX is not the dataset itself; it is the tool that makes it possible for users to manipulate physically realistic DLOs and systematically synthesize wire data at scale.

\begin{figure}[t]
    \centering
    \safeincludegraphics[width=\linewidth]{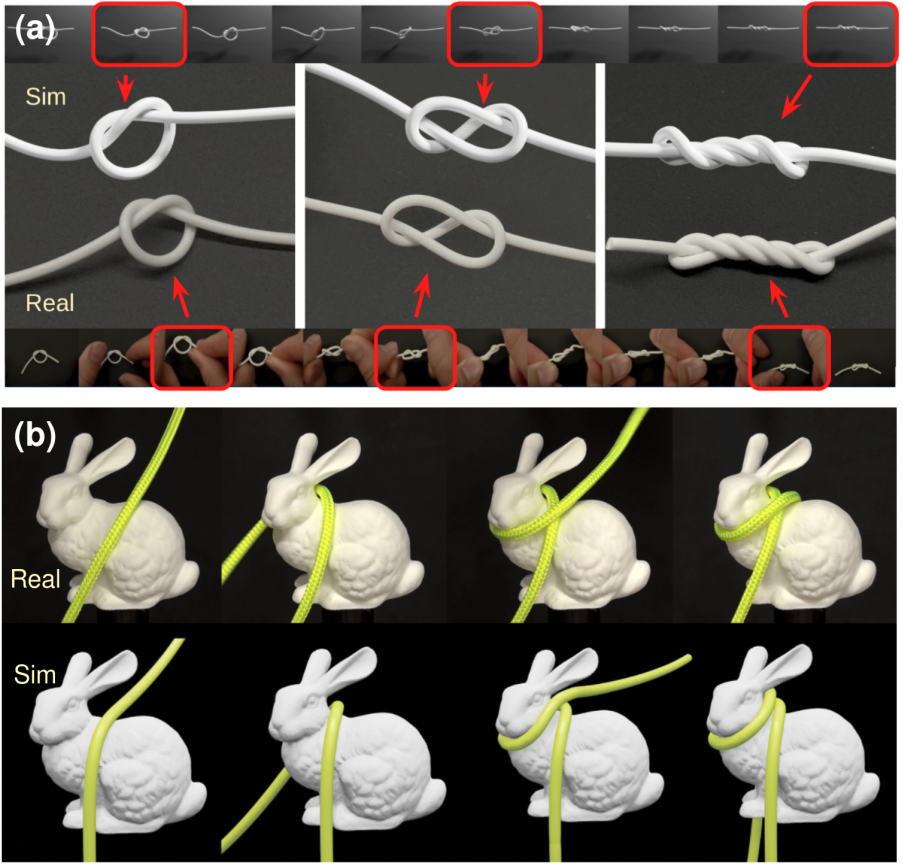}
    \caption{Sim2real comparison for DLOs. (a) We tie a trefoil knot on a wire, apply continuous twists to induce shape transitions, and qualitatively compare the representative configurations between simulation (top) and real experiments (bottom). (b) shows flexible rope (scarf) wrapping around a bunny with multi-point contact, sliding, and self-contact.}
    \label{fig:physics}
\end{figure}

\noindent\textbf{Physics sanity checks for data realism.}
We validates DeformX with two real-world demonstrations that motivate its use for dataset generation. In the first demo, a trefoil knot is twisted and undergoes qualitative shape transitions that match the real wire, showing that the simulator reproduces torsion-bending coupling rather than merely rendering plausible static shapes. In the second demo, a flexible scarf is pulled around a rigid bunny, inducing sliding contact, self-contact, and geometry-conforming draping. As shown in Fig.~\ref{fig:physics}, the simulated configurations remain stable, avoid unrealistic penetration, and qualitatively follow the real trajectories. These demonstrations matter because they directly target the kinds of crossings, wraps, and occlusions that later appear in the synthetic segmentation benchmark.

\begin{figure}[!t]
    \centering
    \includegraphics[width=\linewidth]{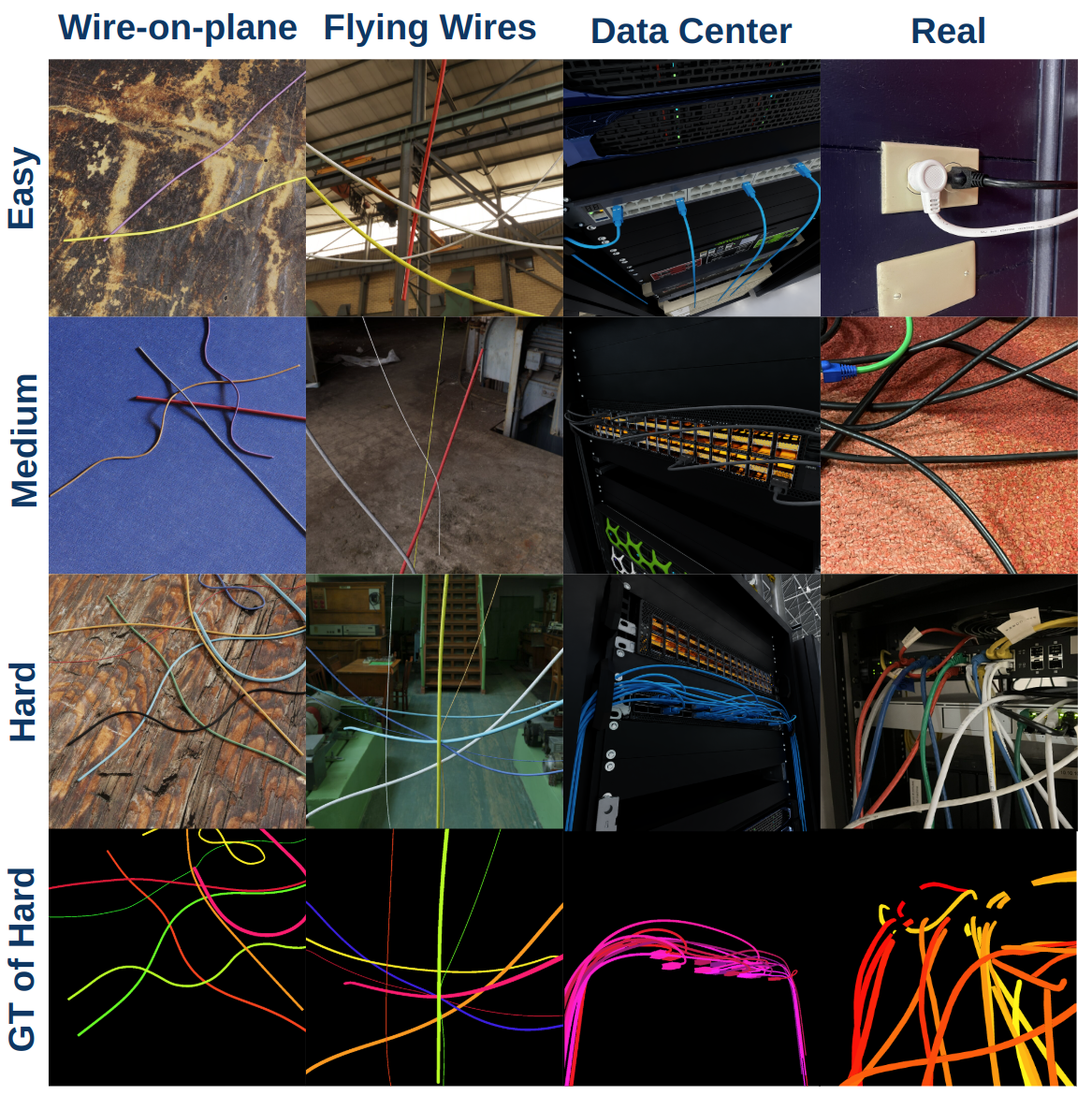}
    \caption{Representative images from three settings across Easy/Medium/Hard, with Hard ground-truth instance masks.}
    \label{fig:dataset}
\end{figure}

\begin{table*}[]
\vspace{5pt}
\centering
\resizebox{\textwidth}{!}{%
\setlength{\tabcolsep}{6pt}
\renewcommand{\arraystretch}{1.15}
\small
\begin{tabular}{lcccccccccc}
\toprule
\multirow{2}{*}{\textbf{Model}} &
\multicolumn{2}{c}{\textbf{Hard (Syn)}} &
\multicolumn{2}{c}{\textbf{Medium (Syn)}} &
\multicolumn{2}{c}{\textbf{Easy (Syn)}} &
\multicolumn{2}{c}{\textbf{Total (Syn)}} &
\multicolumn{2}{c}{\textbf{Total (Real)}} \\
\cmidrule(lr){2-3}\cmidrule(lr){4-5}\cmidrule(lr){6-7}\cmidrule(lr){8-9}\cmidrule(lr){10-11}
& \textbf{F1@75} & \textbf{mAP@75} &
  \textbf{F1@75} & \textbf{mAP@75} &
  \textbf{F1@75} & \textbf{mAP@75} &
  \textbf{F1@75} & \textbf{mAP@75} &
  \textbf{F1@75} & \textbf{mAP@75} \\
\midrule
SAM3 (Base)   & 0.179 & 0.066 & 0.446 & 0.310 & 0.803 & 0.735 & 0.409 & 0.290 & 0.296 & 0.157 \\
SAM3 + LoRA   & 0.225 & 0.102 & 0.512 & 0.404 & 0.850 & 0.816 & 0.465 & 0.365 & 0.314 & 0.173 \\
\midrule
$\Delta$ (LoRA--Base) & +25.7\% & +54.5\% & +14.8\% & +30.3\% & +5.9\%  & +11.0\% & +13.7\%  & +25.7\% & +6.1\%  & +10.2\% \\
\bottomrule
\end{tabular}}
\caption{SAM3 performance with and without LoRA fine-tuning. For each subset, we report dataset-level F1@75 and COCO mAP@75 (higher is better) on the Easy/Medium/Hard tiers, the full synthetic set, and a held-out real test set. The last row reports the percent improvement from LoRA fine-tuning.}
\vspace{-5pt}
\label{tab:sam_result}
\end{table*}

\begin{table}[t]
\centering
\scriptsize
\setlength{\tabcolsep}{3pt}
\renewcommand{\arraystretch}{1.0}
\begin{tabular}{@{}lccccc@{}}
\toprule
Dataset &
\begin{tabular}[c]{@{}c@{}}Physics\\Realism\end{tabular} &
\begin{tabular}[c]{@{}c@{}}Instance\\Label\end{tabular} &
\begin{tabular}[c]{@{}c@{}}CAD\\Support\end{tabular} &
\begin{tabular}[c]{@{}c@{}}Visually-grounded\\Scenes\end{tabular} &
\begin{tabular}[c]{@{}c@{}}Num.\\Images\end{tabular} \\
\midrule
HANDLOOM \cite{handloom} & \No & \No & \No & \No & 30k \\
FASTDLO \cite{fastdlo2022} & \No & \Yes & \No & \No & 32k \\
Fresnillo et al.\ \cite{fresnillo2024} & \Yes & \No & \No & \Yes & 25k \\
Zanella et al.\ \cite{zanella2021} & \No & \No & \No & \No & 28.5k \\
ISCUTE \cite{iscute2024} & \No & \Yes & \Yes & \Yes & 28k \\
\textbf{WireSeg-32k (Ours)} & \Yes & \Yes & \Yes & \Yes & 32k \\
\bottomrule
\end{tabular}
\caption{Binary comparison between existing DLO datasets and our WireSeg-32k dataset. ``Physics realism'' indicates whether data generation uses physics-based simulation of DLO deformation (e.g., gravity, wind, and collisions). ``Instance label'' indicates whether per-object instance masks are provided (trace-only supervision is counted as \No). ``CAD support'' indicates whether DLOs can be represented as free-form meshes (enabling direct CAD import). ``Visually grounded scenes'' indicates whether cables are rendered within realistic, image-based backgrounds rather than on plain or purely synthetic backdrops.}
\label{tab:wire_dataset_binary}
\end{table}

\section{WireSeg-32K Dataset}
\noindent\textbf{Overview.}
WireSeg-32K is designed around three principles: high diversity in wire configurations, a broad spectrum of perceptual difficulty, and scenario categories aligned with realistic deployment settings. Using DeformX, we vary wire count, geometry, material properties, and initial states, then render the resulting scenes in Isaac Sim with randomized viewpoints, lighting, clutter, and textures. Because the labels are derived from simulated DLO identities rather than manual tracing, the pipeline provides consistent per-wire instance masks even for thin, overlapping, and self-occluding structures.

\noindent\textbf{Categories and annotations.}
The dataset contains three scenario categories: \emph{wire-on-plane}, a canonical tabletop setup; \emph{flying wires}, which emphasizes gravity-driven dangling behavior; and \emph{data center}, which places cables in cluttered rack-like environments. In total, WireSeg-32K contains 32,000 rendered images from more than 300 independent simulation runs. Each sample includes RGB, per-wire instance masks, and depth. We also provide a complementary real test set of 300 in-the-wild images with manual instance annotations to evaluate synthetic-to-real transfer.

\noindent\textbf{Difficulty splits and positioning relative to prior work.}
Following the full paper, we stratify the synthetic set by difficulty using SAM3 performance on generated images: samples with AP@75 $<0.3$ are labeled hard, those with AP@75 $>0.6$ are labeled easy, and the remainder are labeled medium. This split is useful because difficulty is not determined only by background appearance; it also reflects wire density, crossings, illumination, and thin-object ambiguity. Compared with prior wire datasets and synthetic pipelines~\cite{handloom,fastdlo2022,zanella2021,fresnillo2024,iscute2024}, WireSeg-32K is designed specifically to combine instance-level labels, physically grounded DLO deformation, photorealistic scene context, and CAD-compatible rendering. That combination is what makes it suitable not only for synthetic training, but also for studying sim-to-real transfer on thin deformable objects.

\section{Vision Benchmark}
To assess the value of WireSeg-32K beyond dataset construction, we next evaluate whether it can provide effective supervision for learning-based wire perception. As a simple benchmark, we test whether training on WireSeg-32K improves wire instance segmentation on both synthetic images and a held-out real-world test set.

We fine-tune Segment Anything Model 3 (SAM3) using lightweight LoRA adapters. Following the full paper, we insert LoRA modules into the attention projections of the vision encoder and mask decoder (\texttt{q\_proj}, \texttt{k\_proj}, and \texttt{v\_proj}), using rank 16, $\alpha=32$, and dropout 0.05, while keeping all remaining SAM3 parameters frozen for parameter-efficient and stable adaptation. Training is performed for 5 epochs on WireSeg-32K with a learning rate of $1\times10^{-5}$ and an effective batch size of 16.

As a foundation-model baseline, we also evaluate off-the-shelf SAM3 in text-prompt mode using the prompt \texttt{cable}. For evaluation, we report dataset-level F1@75, defined as instance F1 with one-to-one matching at IoU $\ge 0.75$, together with COCO mAP@75. We compare performance across the easy, medium, and hard synthetic subsets, the full synthetic benchmark, and the held-out real test set.

Table~\ref{tab:sam_result} shows that LoRA fine-tuning consistently improves performance over the off-the-shelf baseline across all subsets, including real images. The gains are especially clear on harder scenes with crossings, clutter, and thin-object ambiguities, suggesting that the diversity and physical realism of WireSeg-32K provide useful supervision for challenging wire perception.

\section{Conclusion}
We presented WireSeg-32K, a physics-grounded synthetic dataset for wire instance segmentation, together with DeformX, the simulation tool used to generate it. By combining Cosserat rod dynamics, free-form mesh contact, lightweight co-simulation, and mesh-skinned rendering, DeformX enables users to manipulate physically realistic DLOs and synthesize labeled data in realistic scenes. Initial experiments with SAM3 show measurable synthetic-to-real transfer, suggesting that WireSeg-32K is a useful benchmark for perception on thin deformable objects.

\clearpage
{\small
\bibliographystyle{ieeenat_fullname}
\bibliography{main}
}

\end{document}